\documentclass[10pt,twocolumn,letterpaper]{article}

\usepackage{cvpr}      

\usepackage{graphicx}
\usepackage{amsmath}
\usepackage{amssymb}
\usepackage{booktabs}
\usepackage{ulem}

\usepackage[pagebackref,breaklinks,colorlinks]{hyperref}

\usepackage[capitalize]{cleveref}
\crefname{section}{Sec.}{Secs.}
\Crefname{section}{Section}{Sections}
\Crefname{table}{Table}{Tables}
\crefname{table}{Tab.}{Tabs.}

\def\cvprPaperID{17} 
\def\confName{CVPR}
\def\confYear{2022}

\begin{document}

\title{Learning Feature Disentanglement and Dynamic Fusion for \\ Recaptured Image Forensic}


\author{Shuyu Miao, \quad Lin Zheng, \quad Hong Jin\\
Ant Group\\
{\tt\small \{miaoshuyu.msy, zhenglin.zhenglin, jinhong.jh\}@antgroup.com}
}

\maketitle

\begin{abstract}
Image recapture seriously breaks the fairness of artificial intelligent (AI) systems, which deceives the system by recapturing others' images. Most of the existing recapture models can only address a single pattern of recapture (e.g., moiré, edge, artifact, and others) based on the datasets with simulated recaptured images using fixed electronic devices. In this paper, we explicitly redefine image recapture forensic task as four patterns of image recapture recognition, i.e., moiré recapture, edge recapture, artifact recapture, and other recapture. Meanwhile, we propose a novel \textbf{\uline{F}}eature \textbf{\uline{D}}isentanglement and \textbf{\uline{D}}ynamic \textbf{\uline{F}}usion (FDDF) model 
to adaptively learn the most effective recapture feature representation for covering different recapture pattern recognition. Furthermore, we collect a large-scale \textbf{R}eal-scene \textbf{U}niversal \textbf{R}ecapture (RUR) dataset containing various recapture patterns, which is about five times the number of previously published datasets. To the best of our knowledge, we are the first to propose a general model and a general real-scene large-scale dataset for recaptured image forensic. Extensive experiments show that our proposed FDDF can achieve state-of-the-art performance on the RUR dataset.
\end{abstract}

\section{Introduction}
\label{sec:intro}
With the development of deep learning, more and more artificial intelligence applications have been applied in our daily life. Computer vision and image processing have played a great role in this process. Although significant advances have been made, it still meets some challenges of safety and fairness, i.e., one of them is that some people recapture the image as their image for some services. Thus, recaptured image forensic 
\cite{8628746, 10.1007/978-3-319-53465-7_9, anjum2020recapture, sun2020recaptured}
deals with whether an image has been recaptured or not. Effectively solving the problem of image capture will greatly promote the security of artificial intelligence systems.

A great deal of previous research on recaptured image forensic has focused on effectively boosting the performance. Abraham et al. \cite{8628746} designed wavelet decomposition and convolutional neural network to use the normalized intensity values in the image as weights for the frequency strength of moiré pattern. However, this model is only effective for moiré pattern recapture. Yang et al. \cite{10.1007/978-3-319-53465-7_9} proposed a laplacian convolutional neural networks (L-CNN) to improve the noise signal ratio introduced by recapture operations. 
Laplacian filter is beneficial to obtain strong response at the edge of noise, thus this method can well solve the capture pattern with the edge of electronic screen. Anjum et al. 
\cite{anjum2020recapture} presented a novel technique to detect recaptured images with artifacts by exploiting the high-level details present in images and based on that edge profile is obtained. It is obvious that this work facilitates for image recapture with artifact pattern. In conclusion, it is worth noting that all these models concentrate on a single pattern of recapture detection. 

Intuitively, we explicitly redefine the image recapture into four categories, i.e., moiré recapture 
which is a ripple-like pattern, edge recapture caused by the device edge recaptured in the image, artifact recapture generated by the reflection of light, and other recaptures (e.g., the cursor of the mouse appeared in the image).
Thus, designing the universal models to tackle various patterns of recaptures sheds new light on the development of recaptured image forensic and safe image-based AI systems. The greatest challenges of achieving this goal lie in the following issues, i.e., specific feature representation (issue.(1)), reasonable feature fusion
 (issue.(2)), and real-scene universal dataset (issue.(3)).
 
 \begin{figure*}
	\centering
	\includegraphics[width=0.85\linewidth]{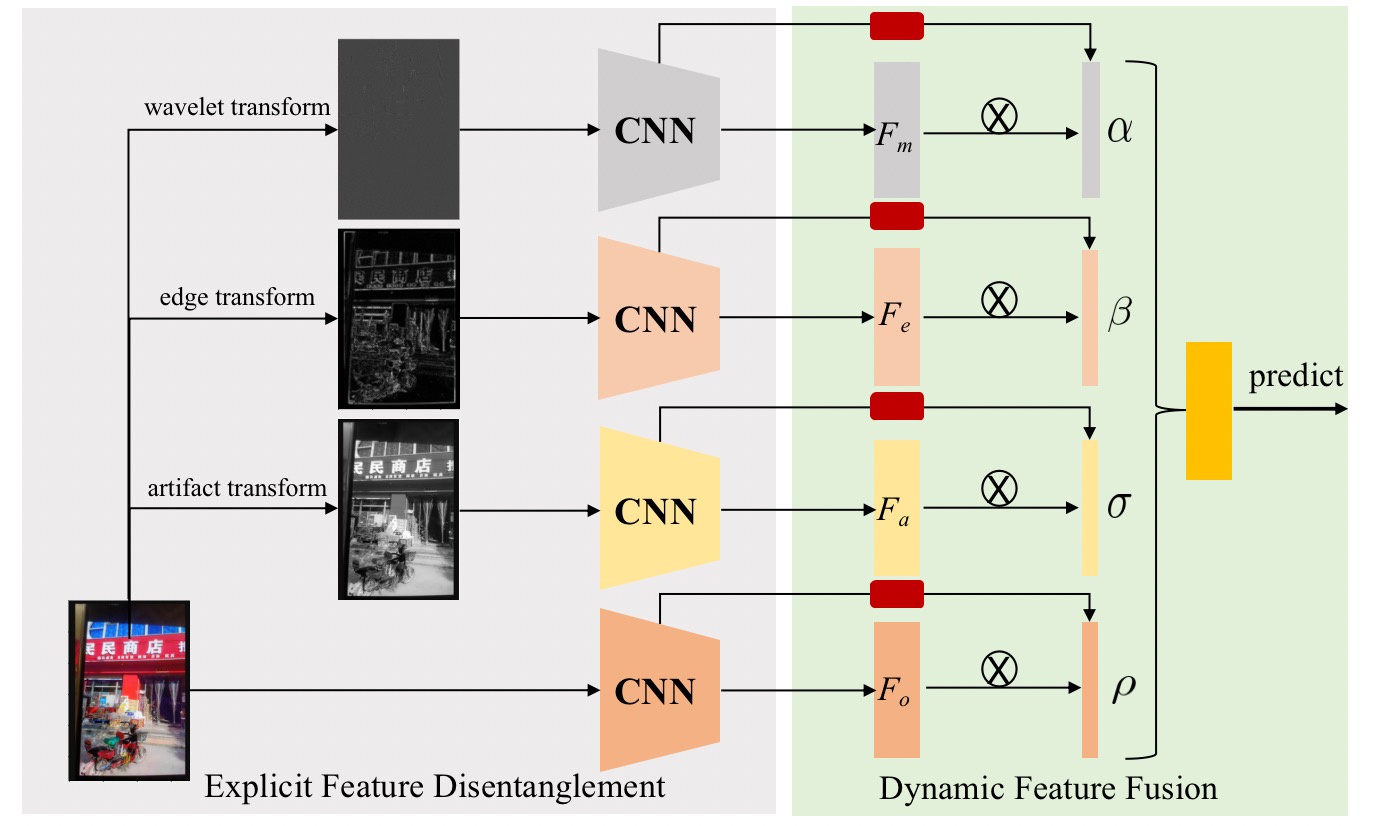}
	\caption{An overview of our novel \textbf{F}eature \textbf{{D}}isentanglement and \textbf{{D}}ynamic \textbf{{F}}usion (FDDF) model.
	}
	\label{fig-model}
\end{figure*}

 To address the above issues, we propose a novel \textit{Feature {{D}}isentanglement and {{D}}ynamic {{F}}usion} model, called  \textbf{FDDF}, to learn the general feature representations of four recapture patterns. In particular, FDDF consists of two main processes, i.e., \textit{explicit feature disentanglement} to address the issue.(1) and  \textit{dynamic feature fusion} to tackle the issue.(2). Firstly, we explicitly disentangle the recapture feature with moiré recapture feature that benefits from wavelet transform, edge recapture feature yielded by laplacian operation, artifact recapture feature by transforming RGB color spaces to YCrCb color spaces for extracting highlight feature, and other recapture feature by squeeze-and-excitation networks \cite{hu2018squeeze}. Secondly, considering that different recaptured images have different recapture feature representation, 
we adopt a weight adaptation module with learnable parameters to allow the model to dynamically learn for itself which feature is most sufficient to fuse the four pattern recapture features for distinguishing whether the image is recaptured or not.  Meanwhile, considering that most of the current publicly available datasets are artificially collected using fixed devices with single recapture pattern, we collect a large-scale \textit{ {R}eal-scene {U}niversal {R}ecapture (RUR)} dataset  containing various recapture patterns to solve the issue.(3), which is about five times the number of previously published datasets. Specifically, our proposed FDDF outperforms other published researches and achieves the best performance on the RUR dataset.

To summarize, our work has the following three main contributions:
\begin{itemize}
\item We propose a novel \textit{Feature {{D}}isentanglement and {{D}}ynamic {{F}}usion (FDDF)}  model to explicitly disentangle the recapture feature and dynamically fuse the feature to address the general image recapture.
\item We collect a large-scale \textit{ {R}eal-scene {U}niversal {R}ecapture (RUR)} dataset  containing various recapture patterns for promoting the development of recaptured image forensic.
\item Extensive experiments show the effectiveness of our proposed FDDF, which achieves the state-of-the-art performance on RUR dataset.
\end{itemize}


\section{Methodology}

\subsection{Overview}
The FDDF model consists of two modules illustrated in Figure \ref{fig-model}, that is, Explicit Feature Disentanglement (EFD) module and Dynamic Feature Fusion (DFF) module. The pipeline of the model is that (1) the image will be fed into four parallel branches with the uniform ResNet18 \cite{he2016deep} as the backbone for generating the explicit four kinds of capture features, (2) the capture features will be fused by learnable module with adaptive weights, (3) and the fused feature is input into the classification layer to determine whether the image is recaptured or not. 

We define the input image as $F_{in}$, and the disentangled features through EFD as $F_m$, $F_e$, $F_a$, and $F_o$, four parallel branches as $Conv_m$, $Conv_e$, $Conv_a$, and $Conv_o$, the learnable DFF as $AdaptLayer$, and the fused feature as $F_{out}$. The pipeline of our proposed FDDF can be formulated as the following expressions.
\begin{equation}
\label{equ_1}
\begin{split}
F_m = Conv_m(F_{in}), F_e = Conv_e(F_{in}) \\ F_a = Conv_a(F_{in}), F_o = Conv_o(F_{in}))
\end{split}
\end{equation} 
\begin{equation}
\label{equ_2}
\begin{split}
F_{out} &= AdaptLayer(F_m, F_e, F_a, F_o) \\
\end{split}
\end{equation} 

\subsection{Explicit Feature Disentanglement}
In order to determine whether an image is recaptured or not, it is usually done by looking at whether the image contains one or more of the recapture features (e.g., moiré, edge, artifact, and others). Thus, Explicit Feature Disentanglement (EFD) module aims to disentangle the recapture features explicitly for extracting  moiré feature, edge feature, artifact recapture, and other features.

\textbf{Moiré recapture pattern.}
Moiré patterns are interference patterns that are
produced due to the overlap of the digital grids of the camera sensor resulting in a high-frequency noise in the image. It is pointed out that wavelet decomposition results in a high response to Moiré noise \cite{8628746}. Inspired by it, we firstly transform the original image $F_{in}$ by discrete wavelet transform to obtain the $LH, HL, HH$ components of the image. Furthermore, we stack the $LH, HL, HH$ as the RGB image, feed it into the backbone of the moiré feature branch, and derive the moiré feature $F_m$. 

\textbf{Edge recapture pattern.}
In many real-scene image recapture scenarios, it is easy to line up the edges of the device into the image. The features of these edges have distinct boundary differentiation points from the normal image content. To extract the edge feature of the recapture, we design the laplacian operation for emphasizing the weights of the edge the device  \cite{wang2007laplacian}. Concretely, we use Gaussian convolution for the original image denoising operation, process the image with grayscale operations,  adopt Laplacian convolution on the grayscale maps, feed it into the backbone of edge feature branch, and yield the edge feature $F_e$.

\textbf{Artifact recapture pattern.}
Shooting at the screen of the device, due to various angles, many times will produce reflections. These reflective features will help in the identification of the recaptured image. For RGB images, each channel contains color information along with luminance information, which is not conducive to processing the specular reflection part. In the YCrCb color space, the luminance and chromaticity information is separated \cite{gorny2005highlight}. When the luminance is lower than 170, it belongs to the diffuse reflection light, the human visual feeling is soft; when the luminance is greater than 200 close to 255, the image presents high light, and belongs to the specular reflection range. Based on this, 
we convert the image from RGB color space to YCrCb color space, feed it into the backbone of the artifact feature branch, and obtain the artifact feature  $F_a$.

\textbf{Other recapture patterns.}
In addition to the above three salient features, there are some other features in recapture image samples that can distinguish whether it is recaptured or not, such as mouse pointer, finger and other cases. For this part of the samples, an attention mechanism squeeze-and-excitation module \cite{hu2018squeeze} is utilized to force the model to learn these distinguishing features, while the attention mechanism can also learn the three salient recapture features mentioned above to some extent.

\subsection{Dynamic Feature Fusion}
For the feature maps obtained from the above four branches, different combinations of features may bring various performance. Because different samples of recapture images have different distinctive features.
To this end, we design a Dynamic Feature Fusion (DFF) module that allows the model to learn the best feature combination scheme adaptively by itself. Thus, the Equation \ref{equ_2} can be further expressed as:
\begin{equation}
\label{equ_3}
F_{out} = \alpha F_m + \beta F_e + \sigma F_a + \rho F_o, \alpha + \beta + \sigma + \rho = 1
\end{equation}

In the specific implementation, $F_m$, $F_e$, $F_a$, and $F_o$ are put into four parallel adaptive layers with the components of \{\textit{convolution layer, batch normalization layer, and relu6 activation layer}\}, operated by concatenation operation, fed into weight redistribution layer by convolution with the output of 4, and processed by softmax function for restricting the four weights to sum to  $1$. 

DFF implements self-learning of the importance of the four features. Concretely speaking, assuming that only moiré features are present in the recaptured image, it is equivalent to that $\alpha$ is set as 1 and $\beta, \sigma,  \rho$ are set as 0. It is just a special case of Equation \ref{equ_3}.

\begin{table}
\centering
\caption{ The statistical table of different datasets.}
\small
\begin{tabular}{c|c|c|c}
\toprule[1.5pt]
Dataset   & Collection & Recaptured Image & Original Image \\ 
\midrule[1.0pt]
LS-D  \cite{agarwal2018diverse}    & artifical  & 145000           & 145000         \\ 
NTU-Rose \cite{cao2010identification} & artificial & 2700             & --             \\ 
ICL  \cite{thongkamwitoon2015image}     & artifical  & 2520             & 1035           \\ 
ASTAR  \cite{gao2010smart}   & real-scene & 1137             & 1094           \\ 
\midrule[1.0pt]
\textbf{RUR (ours)} & \textbf{real-scene} & \textbf{75000}            & \textbf{75000}     \\ \bottomrule[1.5pt]
\end{tabular}
\label{lab_1111}
\end{table}

\section{Experiments}
\subsection{Large-scale Real-scene Universal Recapture (RUR) dataset}
Widely used public image recapture datasets as the Table \ref{lab_1111} include the followings, i.e., LS-D \cite{agarwal2018diverse} that consists of 145000 pairs of original and recaptured images collected by fixed devices artificially, NTU-Rose \cite{cao2010identification} that consists of 2700 recaptured by using digital still cameras and three LCDs artificially, ICL \cite{thongkamwitoon2015image}  that consists of 1035 original images and  2520 images recaptured by using different devices, and ASTAR \cite{gao2010smart}  that consists of 1094 real-scene images and 1137 recaptured images with
real environment background.
The LS-D, NTU-Rose, and ICL datasets collected artificially often differ significantly from the data distribution in real-world scenes. ASTAR collected in the real scene just have few data. To make a step forward, we collect a \textit{ {R}eal-scene {U}niversal {R}ecapture (RUR)} dataset containing various recapture patterns with 75000 recaptured images and 75000 normal images. 

\begin{table}
\centering
\caption{ The experiments of different models on  RUR datasets.}
\begin{tabular}{cccc}
\toprule[1.5pt]
Model & dataset        & Precision & Recall \\
\midrule[1.0pt]
Abraham et al. \cite{8628746}  & RUR & 95.6\%    & 92.3\% \\
Yang et al.  \cite{10.1007/978-3-319-53465-7_9} & RUR & 96.2\%    & 93.6\% \\
Anjum et al. \cite{anjum2020recapture} & RUR & 94.2\%    & 91.8\% \\
Sun et al. \cite{sun2020recaptured}   & RUR & 96.8\%    & 94.1\% \\
\midrule[1.0pt]
\textbf{FDDF (ours)}   & \textbf{RUR}  & \textbf{98.1}\%    & \textbf{95.3}\% \\
\bottomrule[1.5pt]
\end{tabular}
\label{table_2}
\end{table}

\begin{table}
\centering
\caption{ The ablation studies of every component in FDDF.}
\begin{tabular}{cccc}
\toprule[1.5pt]
Components & dataset        & Precision & Recall \\
\midrule[1.0pt]
Without Moiré  & RUR & 96.6\%    & 94.2\% \\
Without Edge & RUR & 96.9\%    & 94.6\% \\
Without Artifact & RUR & 97.8\%    & 94.9\% \\
Without Others  & RUR & 97.4\%    & 94.5\% \\
\midrule[1.0pt]
FDDF (all)   & RUR  & 98.1\%    & 95.3\% \\
\bottomrule[1.5pt]
\end{tabular}
\label{table_3}
\end{table}

\subsection{Overall Performance}
We evaluate the performance of our proposed FDDF on the proposed RUR dataset. The related experimental results are
shown in Table \ref{table_2}. From the table, we can see that our proposed FDDF can obtain the best performance on the RUR in precision and recall. FDDF outperforms \cite{8628746} in the precision and recall benchmark by 2.5\% and 3.0\%, \cite{10.1007/978-3-319-53465-7_9} by 1.9\% and 1.7\%, \cite{anjum2020recapture} by 3.9\% and 3.5\%, and \cite{sun2020recaptured} by 1.3\% and 1.2\%, respectively. It demonstrates the superiority of our approach FDDF.

\subsection{Ablation Study}
To verify the reasonability and reliability of our proposed FDDF, we perform the related ablation studies as the Table \ref{table_3}. `Without xxx' means to remove the corresponding feature branch in the FDDF. It can be observed from the table when without the moiré branch, the precision and recall drop by 1.5\% and  1.1\%; when without the edge branch, the precision and recall drop by 1.2\% and  0.7\%; when without the artifact branch, the precision and recall drop by 0.3\% and  0.4\%; and when without edge branch, the precision and recall drop by 0.7\% and  0.8\%. 
The experimental results show that each feature branch is critical to the performance of the proposed FDDF. 
 
\section{Conclusion}
To break the threat of recaptured images to artificial intelligence systems, we present a novel Feature Disentanglement and Dynamic Fusion (FDDF) model to explicitly disentangle the recapture feature and dynamically fuse the feature to detect image recapture with various patterns. Meanwhile, we collect a large-scale Real-scene Universal Recapture (RUR) dataset to promote the development of recaptured image forensic. We hope our work can shed new light on this task and AI system safety.

{\small
\bibliographystyle{ieee_fullname}
\bibliography{reference}
}

\end{document}